\documentclass[11pt]{article}

\usepackage[final]{acl}

\usepackage{times}
\usepackage{latexsym}

\usepackage[T1]{fontenc}

\usepackage[utf8]{inputenc}

\usepackage{microtype}

\usepackage{inconsolata}

\usepackage{graphicx}
\usepackage{booktabs}
\usepackage{multirow}
\usepackage{tabularx}
\usepackage{listings}
\usepackage{booktabs}
\usepackage{pifont}
\usepackage{multirow}
\usepackage{amsmath}
\usepackage{dsfont}
\usepackage[most]{tcolorbox}
\tcbuselibrary{breakable}
\usepackage{amssymb}
\usepackage{rotating}
\usepackage{enumitem}
\usepackage[dvipsnames]{xcolor}

\definecolor{deepskyblue}{HTML}{00bfff}
\definecolor{darkcyan}{HTML}{008b8b}
\definecolor{deepgreen}{rgb}{0.0, 0.5, 0.0}
\definecolor{light_blue}{RGB}{243, 247, 253}
\definecolor{light_pink}{RGB}{252, 241, 242}
\definecolor{light_yellow}{RGB}{255, 255, 233}

\newcommand{\xmark}{\textcolor{red}{\textbf{\ding{55}}}}
\newcommand{\cmark}{\textcolor{deepgreen}{\textbf{\ding{51}}}}

\usepackage{tikz}
\usepackage{amsmath}
\usepackage{amssymb}
\usetikzlibrary{positioning, arrows.meta, calc, fit, backgrounds, shapes.geometric}

\title{Long-Context Reasoning Through Proxy-Based Chain-of-Thought Tuning}
\author{Miao Li, Irina Saparina, Alexander Gurung, Mirella Lapata\\
        School of Informatics,\\  University of Edinburgh\\
         \texttt{\{miao.li@, alex.gurung@, i.saparina@sms.,   mlap@inf.\}ed.ac.uk} 
         }

\begin{document}
\maketitle
\begin{abstract}

Recent large language models support inputs of up to 10 million tokens, yet they perform poorly on long-context tasks that require complex reasoning. 
Such tasks can be solved using only a subset of the input --- a proxy context --- rather than the full sequence. Despite sharing the same underlying reasoning process, models exhibit a significant performance disparity between proxy and full contexts. To improve long-context reasoning, we propose ProxyCoT, a novel training framework that transfers reasoning capabilities from short proxy contexts to full long contexts.  Specifically, we first obtain high-quality chain-of-thought reasoning traces on proxy contexts through reinforcement learning or distillation from a larger teacher model, and then ground the generated traces in full long contexts with supervised fine-tuning. Experiments across different datasets demonstrate that ProxyCoT consistently outperforms strong baselines with reduced computational overhead. Furthermore, models trained with ProxyCoT generalize their long-context reasoning capabilities to out-of-domain tasks.\footnote{Our code, data, and models are available at \url{https://github.com/oaimli/ProxyCoT}.}
\end{abstract}

\section{Introduction}

Large language models (LLMs) have been developed with increasingly expansive context windows, now reaching lengths of up to 10 million tokens~\citep{gemini25_2025, llama4_2025, qwen25_1m_2025}. These models promise advancements for long-context tasks that demand complex reasoning, such as synthesizing insights from multiple medical reports or addressing analytical questions spanning several financial documents. Successfully performing these tasks requires LLMs not only to locate relevant information within extensive inputs but also to reason effectively over the extracted knowledge to produce correct responses.


\begin{figure}[t]
\centering
\includegraphics[width=0.5\textwidth, trim=12 6 20 35, clip]{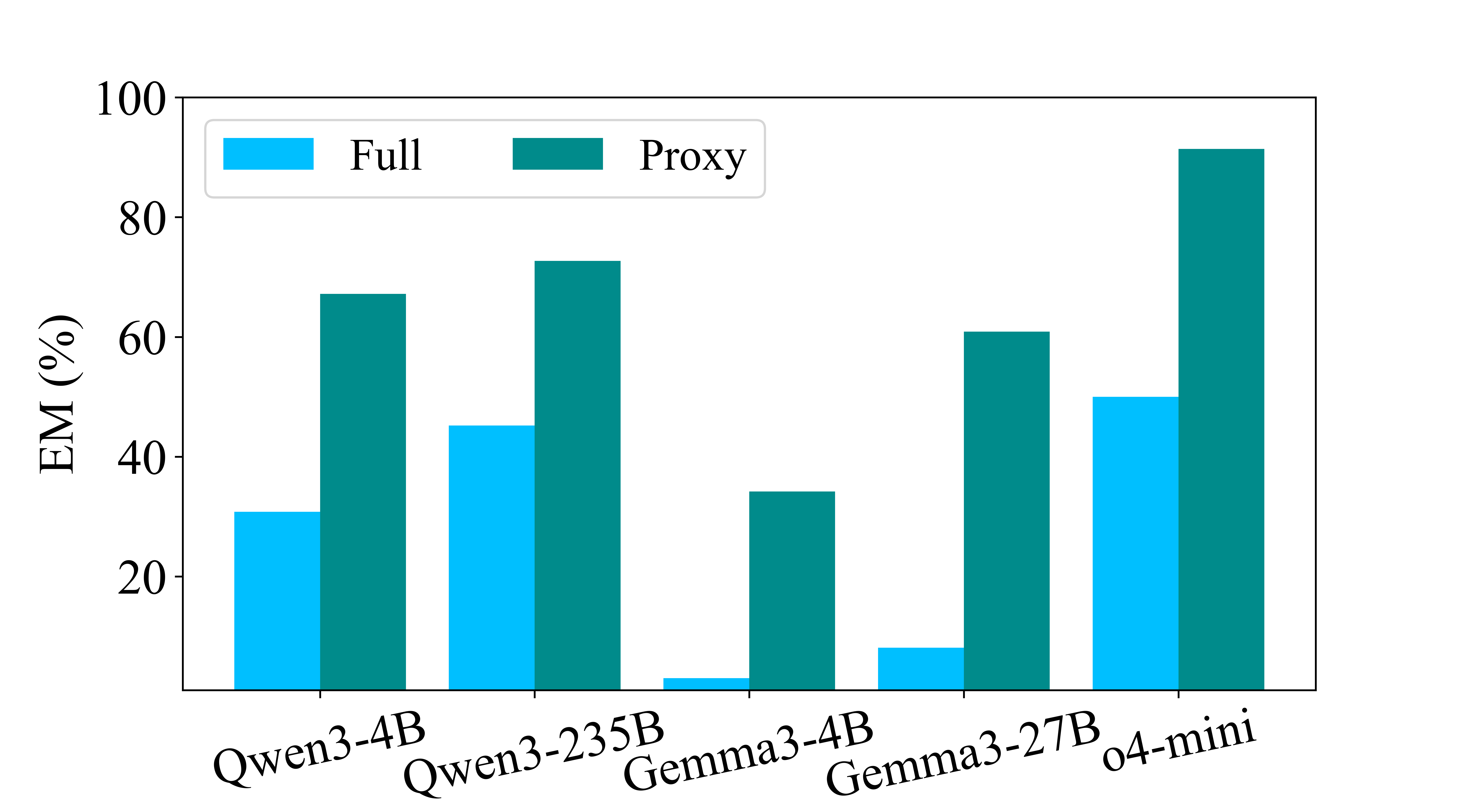}
\caption{The disparity of model performance in the zero-shot setting on SciTrek \cite{scitrek_2025}  when showing \textit{full long contexts} vs \textit{short proxy contexts} in terms of exact match. The full context includes 128K tokens, while there are only around 650 tokens on average in a proxy context. Models work better on proxy contexts, while the proxy context requires the same reasoning process as their corresponding full long context.}
\label{fig:gap_scitrek}
\end{figure}

To enhance the reasoning capabilities of LLMs, prior work has primarily relied on the chain-of-thought distillation~\citep{symbolic_cot_distillation_2023, reasoning_teacher_2023} and reinforcement learning~\citep{deepseekai2025deepseekr1} to elicit visible, step-by-step reasoning traces. These approaches have been successful on short-context tasks, but exhibit notable limitations when applied to long-context settings. Chain-of-thought distillation, for instance, depends on the high-quality reasoning traces from a teacher model, which is typically large and therefore slow and costly to query~\citep{symbolic_cot_distillation_2023, reasoning_teacher_2023, deepseekai2025deepseekr1}.\footnote{Closed-source models from OpenAI and Google do not even provide access to their reasoning traces.} Moreover, even strong teacher models may produce unreliable traces on complex long-context tasks. As an illustration, on SciTrek, a recently released long-context question answering benchmark over (full-text) scientific articles, the best performing open-source model only achieves 48.8\% exact match~\citep{scitrek_2025}. Reinforcement learning with policy-gradient methods, often used when no suitable teachers are available, is also challenging to scale to long contexts because it requires extensive sampling, making training slow and computationally expensive.


However, there are indications that much of the computational cost of long-context processing is unnecessary: in many long-context tasks, only a small fraction of the input provides the evidence needed for the correct output.  For instance, in multi-hop QA, such as HotpotQA~\citep{hotpotqa_2018}, systems may retrieve entire articles, yet the answer typically depends on only a handful of relevant sentences.  We refer to such subsets as \textit{proxy contexts}: compact snippets that contain sufficient information to derive the correct answer. We hypothesize that the underlying reasoning should be \textit{invariant} to the choice of context representation, i.e., a model should follow the same reasoning steps, whether it is conditioned on the full long context or on the corresponding proxy context.

Although the underlying information and required reasoning are the same, our experiments reveal a substantial performance gap between full and proxy contexts. As shown in Figure~\ref{fig:gap_scitrek}, LLMs across scales and model families perform markedly better when conditioned on proxy contexts. When given the full context, \citet{scitrek_2025} report that models often produce plausible \emph{high-level} reasoning structures, yet hallucinate the \emph{specific} facts needed to execute those steps correctly. In contrast, we find that the same models achieve much higher accuracy on each reasoning step when given proxy contexts.  This suggests that LLMs struggle to correctly ground their reasoning in the relevant evidence in long inputs.

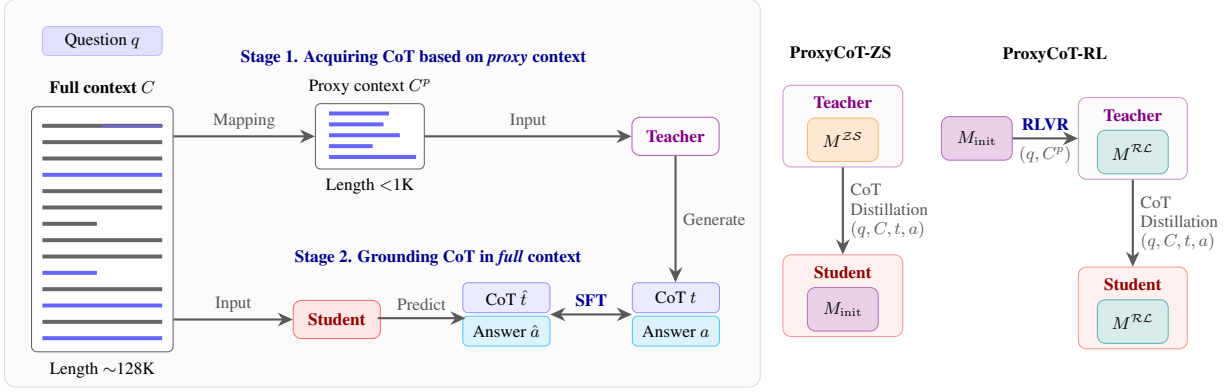
\begin{figure*}[t]
\centering
\begin{tikzpicture}[
  font=\small,
  scale=0.72, transform shape,>={Stealth[length=2mm,width=1.8mm]},
  teacher/.style={
    draw=violet!60, fill=violet!2, rounded corners=3pt,
    minimum width=1.6cm, minimum height=0.7cm,
    text=violet!55!black, font=\bfseries\small
  },
  student/.style={
    draw=red!55,    fill=red!10,    rounded corners=3pt,
    minimum width=1.6cm, minimum height=0.7cm,
    text=red!55!black, font=\bfseries\small
  },
  cotbox/.style={
    draw=blue!40,   fill=blue!8,    rounded corners=2pt,
    minimum width=1.6cm, minimum height=0.55cm,
    font=\small
  },
  ansbox/.style={
    draw=cyan!55,   fill=cyan!12,   rounded corners=2pt,
    minimum width=1.6cm, minimum height=0.55cm,
    font=\small
  },
  modelA/.style={
    draw=orange!55, fill=orange!18, rounded corners=3pt,
    minimum width=1.3cm, minimum height=0.8cm, font=\small
  },
  modelB/.style={
    draw=teal!60,   fill=teal!15,   rounded corners=3pt,
    minimum width=1.3cm, minimum height=0.8cm, font=\small
  },
  modelC/.style={
    draw=violet!55, fill=violet!18, rounded corners=3pt,
    minimum width=1.3cm, minimum height=0.8cm, font=\small
  },
  stagebg/.style={
    draw=blue!15, fill=blue!4, rounded corners=4pt, inner sep=6pt
  },
  panelbg/.style={
    draw=gray!25, fill=gray!3, rounded corners=4pt, inner sep=8pt
  },
  arr/.style={->, thick, gray!70!black},
  doublearr/.style={<->, thick, gray!70!black}
]


\node[draw=black!60, fill=white, minimum width=2.6cm, minimum height=4.5cm,
      rounded corners=2pt] (fullctx) at (0,0) {};
\node[anchor=south, font=\small\bfseries]
  at ([yshift=0.1cm]fullctx.north) {Full context $C$};
\node[draw=blue!25, fill=blue!12, rounded corners=2pt,
      minimum width=2.2cm, minimum height=0.5cm, font=\small]
  (question) at ([yshift=1.2cm]fullctx.north) {Question $q$};
\node[anchor=north, font=\small, black]
  at ([yshift=-0.05cm]fullctx.south) {Length $\sim$128K};

\foreach \y/\w/\c in {
  1.9/2.2/black, 1.6/2.2/black, 1.3/2.2/black, 1.0/2.2/blue,
  0.7/2.2/black, 0.4/2.2/black, 0.1/1/black, -0.2/2.2/black,
  -0.5/2.2/black, -0.8/1/blue, -1.1/2.2/black, -1.4/2.2/blue,
  -1.7/2.2/black, -2.0/2.2/blue} {
  \draw[\c!60, line width=1.5pt]
  ([xshift=-1.1cm, yshift=\y cm]fullctx.center) -- ++(\w cm, 0);
  \draw[black!60, line width=0.6pt]
  ([xshift=-1.1cm, yshift=1.9 cm]fullctx.center) -- ++(1.1cm, 0);
\draw[blue!60, line width=0.6pt]
  ([xshift=0cm, yshift=1.9 cm]fullctx.center) -- ++(1.1cm, 0);
}

\node[draw=black!60, fill=white, minimum width=2.0cm, minimum height=1.2cm,
      rounded corners=2pt, right=2.6cm of fullctx.north east, yshift=-0.55cm]
      (proxy) {};
\node[anchor=south, font=\small] at ([yshift=0.02cm]proxy.north) {Proxy context $C^p$};
\foreach \y/\w in {0.42/1.1, 0.22/1.0, 0.02/1.3, -0.18/0.8, -0.38/1.6} {
  \draw[blue!60, line width=1.5pt]
    ([xshift=-0.75cm, yshift=\y cm]proxy.center) -- ++(\w cm, 0);
}
\node[anchor=north, font=\small, black]
  at ([yshift=-0.02cm]proxy.south) {Length $<$1K};

\node[teacher, right=3.8cm of proxy] (teacher1) {\textcolor{violet}{Teacher}};

\node[anchor=south, font=\small\bfseries, text=blue!55!black]
  at ([xshift=0.8cm, yshift=0.55cm]proxy.north)
  (stage1title) {Stage 1. Acquiring CoT based on \emph{proxy} context};

\draw[arr] (fullctx.east |- proxy.west) -- node[above, font=\small] {Mapping} (proxy.west);

  \begin{pgfonlayer}{background}

  \node[stagebg, fit=(stage1title)(proxy)(teacher1), inner sep=5pt] (stage1bg) {};
\end{pgfonlayer}

\node[student] (student) at ([xshift=3.0cm, yshift=.6cm]fullctx.south east) {Student};

\node[cotbox, right=0.9cm of student, xshift=.6cm, yshift=0.4cm] (cothat) {CoT $\hat{t}$};
\node[ansbox, below=0.05cm of cothat] (anshat) {Answer $\hat{a}$};

\node[cotbox, right=1.5cm of cothat] (cot) {CoT $t$};
\node[ansbox, below=0.05cm of cot] (ans) {Answer $a$};

\node[anchor=south west, font=\small\bfseries, text=blue!55!black]
  at ([xshift=-0.1cm, yshift=0.5cm]student.north west)
  (stage2title) {Stage 2. Grounding CoT in \emph{full} context};



\draw[arr] (proxy.east) -- node[above, font=\small] {Input} (teacher1.west |- proxy.east);

  \draw[arr] ([yshift=-5pt]$(cothat.west)!0.5!(student.west)$) --
  node[above,font=\small] {Predict}
  ([yshift=-5pt]$(cothat.east)!0.5!(student.east)$);
     
\draw[doublearr] ($(cothat.east)!0.5!(anshat.east)$) --
  node[above, font=\small\bfseries, text=blue!60!black] {SFT}
  ($(cot.west)!0.5!(ans.west)$);

\draw[arr] (teacher1.south |- teacher1.south) -- node[right,
  font=\small, pos=0.5] {Generate} (teacher1.south |- cot.north);


\draw[arr] (fullctx.east |- student.west) -- node[above, font=\small] {Input} (student.west);

\begin{pgfonlayer}{background}
  \node[stagebg, fit=(stage2title)(student)(cothat)(anshat)(cot)(ans), inner sep=5pt] (stage2bg) {};
\end{pgfonlayer}

\begin{pgfonlayer}{background}
  \node[panelbg, fit=(question)(fullctx)(stage1bg)(stage2bg), inner sep=10pt] (leftpanel) {};
\end{pgfonlayer}


\node[anchor=south west, font=\small\bfseries,
      right=.4cm of leftpanel.north east, yshift=-1cm] (zs_title) {ProxyCoT-ZS};

\node[draw=violet!40, fill=violet!2, rounded corners=3pt,
      minimum width=2.2cm, minimum height=1.5cm,
      below=0.3cm of zs_title.south west, anchor=north west] (zs_teacher_box) {};
\node[anchor=north, font=\small\bfseries, text=violet!55!black]
  at ([yshift=-0.08cm]zs_teacher_box.north) {\textcolor{violet}{Teacher}};
\node[modelA] at ([yshift=-0.25cm]zs_teacher_box.center) (zs_model_teacher)
  {$M^{\mathcal{ZS}}$};

\node[draw=red!40, fill=red!6, rounded corners=3pt,
      minimum width=2.2cm, minimum height=1.5cm,
      below=1.6cm of zs_teacher_box] (zs_student_box) {};
\node[anchor=north, font=\small\bfseries, text=red!55!black]
  at ([yshift=-0.08cm]zs_student_box.north) {Student};
\node[modelC] at ([yshift=-0.25cm]zs_student_box.center) (zs_model_student)
  {$M_{\mathrm{init}}$};

\draw[arr] (zs_teacher_box.south) -- node[right, font=\small, align=left]
  {CoT \\Distillation\\ $(q,C,t,a)$} (zs_student_box.north);

\begin{pgfonlayer}{background}
  \node[fit=(zs_title)(zs_teacher_box)(zs_student_box), inner sep=10pt] (midpanel) {};
\end{pgfonlayer}


\node[anchor=south west, font=\small\bfseries,
      right=.2cm of midpanel.north east, yshift=-0.8cm] (rl_title) {\hspace{1cm}ProxyCoT-RL};

\node[modelC, below=0.85cm of rl_title.south west, anchor=north west] (rl_init)
  {$M_{\mathrm{init}}$};

\node[draw=violet!40, fill=violet!2, rounded corners=3pt,
      minimum width=2.0cm, minimum height=1.5cm,
      right=1.2cm of rl_init] (rl_teacher_box) {};
\node[anchor=north, font=\small\bfseries, text=violet!55!black]
  at ([yshift=-0.08cm]rl_teacher_box.north) {\textcolor{violet}{Teacher}};
\node[modelB] at ([yshift=-0.25cm]rl_teacher_box.center) (rl_model_teacher)
  {$M^{\mathcal{RL}}$};

\draw[arr] (rl_init.east) -- node[above, font=\small\bfseries, text=blue!60!black] {RLVR}
  node[below, font=\small] {$(q,C^p)$} (rl_teacher_box.west);

\node[draw=red!40, fill=red!6, rounded corners=3pt,
      minimum width=2.0cm, minimum height=1.5cm,
      below=1.6cm of rl_teacher_box] (rl_student_box) {};
\node[anchor=north, font=\small\bfseries, text=red!55!black]
  at ([yshift=-0.08cm]rl_student_box.north) {Student};
\node[modelB] at ([yshift=-0.25cm]rl_student_box.center) (rl_model_student)
  {$M^{\mathcal{RL}}$};

\draw[arr] (rl_teacher_box.south) -- node[right, font=\small, align=left]
  {CoT \\Distillation\\ $(q,C,t,a)$} (rl_student_box.north);

\begin{pgfonlayer}{background}
  \node[ fit=(rl_title)(rl_init)(rl_teacher_box)(rl_student_box), inner sep=15pt] (rightpanel) {};
\end{pgfonlayer}

\end{tikzpicture} 

\caption{General two-stage pipeline of ProxyCoT (left), and two instantiations (right): ProxyCoT-ZS and ProxyCoT-RL. 
Given a target model $M_{\text{init}}$, ProxyCoT-ZS employs a large off-the-shelf model $M^\mathcal{ZS}$ as the teacher to generate CoTs from proxy contexts, and then fine-tunes $M_{\text{init}}$ as the student to generate CoTs on corresponding long contexts. ProxyCoT-RL first optimizes $M_{\text{init}}$ using RLVR to obtain CoTs on proxy contexts, and subsequently fine-tunes the RL-optimized model $M^\mathcal{RL}$ as the student to ground these CoTs in corresponding long contexts.}
\label{fig:proxy_pipeline}
\end{figure*}

Because performance is often substantially higher on proxy contexts and reinforcement learning on them is far less computationally expensive, this motivates us to use the proxy contexts as a means for improving long-context reasoning. 
Figure~\ref{fig:proxy_pipeline} provides an overview of our training framework. We first obtain chain-of-thought reasoning traces (CoTs) based on proxy contexts (e.g., in SciTrek, metadata can serve as a proxy for the full-text articles). These traces can be obtained by reinforcement learning with verifiable rewards or by sampling from a larger teacher model.  We then perform CoT distillation via supervised fine-tuning (SFT), training the target model to reproduce the proxy-derived reasoning traces \textit{when given the full long contexts}.

This two-stage procedure first teaches the model to reason in a computationally efficient setting and then transfers that reasoning behaviour to long inputs. Compared to reinforcement learning directly on full contexts, our framework significantly reduces training cost and avoids requiring teacher-generated traces over long contexts. We summarize our contributions as follows:

\begin{itemize}[noitemsep, topsep=0pt]
\item We introduce and formalize \textit{proxy contexts} for long-context tasks, revealing a significant performance disparity between short proxy and full long contexts. 

\item We propose ProxyCoT, a novel training framework that leverages the short proxy contexts to acquire high-quality chain-of-thought reasoning traces, which in turn are used to enhance reasoning over full long contexts.
\item Through extensive experiments across multiple models and datasets, we demonstrate that ProxyCoT consistently outperforms strong baselines while generating shorter reasoning traces, and generalizes to out-of-domain long-context reasoning tasks.
 
\end{itemize}

\section{Related Work}

\paragraph{Reasoning in Language Models}


Reinforcement learning and chain-of-thought distillation from teacher models are widely-used approaches for improving language model reasoning~\citep{llm_posttraining_2025}. DeepSeek-R1~\citep{deepseekai2025deepseekr1} showed that reasoning abilities can be developed through pure reinforcement learning without requiring supervised fine-tuning as a first step. This is especially useful for frontier models where `teachers' may not exist to provide reasoning traces. However, this often comes at a high computational cost, even for short-context tasks like mathematics. 

DeepSeek-R1 has inspired many subsequent efforts in training reasoning models~\citep{yang2025qwen3, bakouch2025smollm3, mistralai2025magistral}. For smaller or non-frontier models, DeepSeek further showed that distilling reasoning patterns from larger models into smaller ones via SFT on reasoning traces~\citep{symbolic_cot_distillation_2023, reasoning_teacher_2023} outperforms applying reinforcement learning directly.  It has since become standard practice to build SFT reasoning datasets through collecting traces from teacher models~\citep{guha2025openthoughts, openr1, li2025naturalthoughts}.

However, both approaches face significant challenges when applied to long-context reasoning. Reinforcement learning becomes prohibitively expensive as context length increases due to extensive sampling over long sequences during training. CoT distillation avoids sampling costs, but still requires querying large teacher models on full long contexts to generate reasoning traces, which is both time-consuming and computationally intensive. Moreover, even strong teacher models may fail to produce reliable traces for difficult long-context tasks. Our approach addresses these limitations by using proxy contexts to generate intermediate training signals, enabling efficient training without requiring long-context inference from teacher models.



\paragraph{Long-context Language Models}

The ability to fully utilize long sequences has been a long-standing challenge for language models~\citep{long_context_survey_2025}. 
A core difficulty is representing token positions over long sequences. Rotary Position Embeddings (RoPE; \citealt{roformer_su2024}) replace absolute positional embeddings with rotational transformations, and subsequent extensions (e.g., YaRN) rescale positional frequencies to support longer contexts without full retraining~\citep{peng2024yarn}. Another bottleneck is the quadratic cost of Transformer attention. Sparse-attention models (e.g., Longformer) reduce computation by restricting attention patterns to selected entries of the full matrix, improving both prefilling and inference efficiency~\citep{beltagy2020longformerlongdocumenttransformer, jiang2024minference, fu2024moamixturesparseattention}. Many contemporary long-context LLMs interleave sparse and full-attention layers to balance efficiency and quality~\citep{llama3_2024, qwen3_technical_report_2025, gemma3_technical_report_2025}. 

Many modern LLMs also incorporate long-context-specific training data and post-training procedures. For example, Qwen2.5-1M~\citep{qwen25_1m_2025} and Qwen3~\citep{qwen3_technical_report_2025} use synthetic long-context data during pre-training and multi-stage supervised fine-tuning tailored to long contexts. Similarly, OLMo 3~\citep{olmo2025olmo} includes curated long-context data and synthetic aggregation-style tasks. However, due to the expense and evaluation challenges of long-context supervision, such training often targets generic long-context understanding rather than eliciting faithful, step-by-step reasoning on downstream tasks.

In this work, we \emph{reformulate} long-context tasks into a setting that enables collecting high-quality reasoning traces (via proxy contexts), and then train models to reproduce these traces when conditioned on the original long context.

\section{ProxyCoT Training}

This section introduces ProxyCoT, our two-stage training framework designed to enhance long-context reasoning in question answering. Given proxy contexts containing the minimal information required to answer each question, ProxyCoT operates in a teacher-student paradigm with two stages, shown in Figure~\ref{fig:proxy_pipeline}. In Stage~1, a teacher model generates high-quality reasoning traces over  \emph{proxy} contexts. In Stage~2, these reasoning traces are used to fine-tune a student model on the corresponding \emph{long} contexts via chain-of-thought distillation. As the reasoning traces can be obtained from a large off-the-shelf model or based on reinforcement learning, ProxyCoT has two variants: ProxyCoT-ZS and ProxyCoT-RL (Figure~\ref{fig:proxy_pipeline} right). 


\subsection{Acquiring CoTs on Short Proxy Contexts}
\label{sec:approach_rlvr}

For any long-context question answering task, a \emph{proxy context} $C^{p}$ denotes a compact version of the long input $C$ that preserves answerability. CoTs over proxy contexts should transfer to the corresponding full long contexts. Formally, we denote each example as a question--context pair $(q, C)$ with ground-truth answer~$a$. Each context has a corresponding proxy, $C^{p}$, which is substantially shorter ($|C^{p}| \ll |C|$) while containing sufficient information to answer the question. Our goal is to obtain a dataset $\mathcal{D} = \{(q_i, C_i^p, t_i, a_i)\}$ of reasoning traces $t$ generated conditioned on proxy contexts. 

\paragraph{Large Teacher Generation} 
We query a large off-the-shelf teacher model~$M^{\mathcal{ZS}}$ to generate reasoning traces conditioned on proxy contexts: $t \sim p_{\phi}(t \mid q, C^{p})$, where $\phi$ denotes the teacher model parameters. We retain only traces that yield correct answers, resulting in high-quality demonstrations. The teacher estimates the distribution over reasoning traces given the question and proxy context. Compared to generating traces over full long contexts, inference on shorter proxies is substantially faster and more cost-effective.

\paragraph{Reinforced Self-exploration} 
If a capable off-the-shelf large teacher model is unavailable (e.g., it is too expensive to run or performs poorly on our specific task), we train the target model $M_{\text{init}}$ directly on proxy contexts using reinforcement learning with verifiable rewards (RLVR). RLVR optimizes $p_{\theta}(t \mid q, C^{p})$ to maximize the probability of generating traces that lead to correct answers. After training, the resulting model $M^{\mathcal{RL}}$ with parameters $\theta_{\text{RL}}$ estimates $p_{\theta_{\text{RL}}}(t \mid q, C^{p})$, from which we sample reasoning traces for the dataset $\mathcal{D}$. This approach enables the model to learn task-specific reasoning structure while focusing exclusively on the necessary information in proxy contexts.

\subsection{Grounding CoTs in Full Long Contexts}
\label{sec:approach_sft}

Stage~2 transfers the reasoning patterns learned through short proxy contexts to full long contexts. The student model must learn to reproduce reasoning traces while conditioning on the full context $C$ rather than the proxy $C^{p}$.  We achieve this through supervised fine-tuning (SFT) on reasoning traces conditioned on $(q, C)$. Crucially, the two ProxyCoT variants (illustrated in the right part of Figure~\ref{fig:proxy_pipeline}) differ in their Stage~2 objectives:

\paragraph{ProxyCoT-ZS} The off-the-shelf teacher model $M^{\mathcal{ZS}}$ generates traces from proxy inputs: $t \sim p_{\phi}(t \mid q, C^{p})$. We then use the target model $M_{\text{init}}$ as the student to reproduce the teacher's reasoning structure while grounding it in the full long context. We fine-tune $M_{\text{init}}$ by minimizing:
\begin{equation}
\mathcal{L}_{\text{SFT}} = -\mathbb{E}_{(q,C,C^p,t)} \left[ \log p_{\theta}(t \mid q, C) \right]
\end{equation}

\paragraph{ProxyCoT-RL} The RL-optimized model $M^{\mathcal{RL}}$ acts as the teacher generating reasoning traces from proxy inputs: $t \sim p_{\theta_{\text{RL}}}(t \mid q, C^{p})$. We then \emph{further train} $M^{\mathcal{RL}}$ as the student to minimize: 
\begin{equation}
\mathcal{L}_{\text{SFT}} = -\mathbb{E}_{(q,C,C^p,t)} \left[ \log p_{\theta_{\text{RL}}}(t \mid q, C) \right]
\end{equation}
This continues training from the RL checkpoint, teaching the model to apply its learned reasoning patterns to full long contexts. As shown in our ablations (\autoref{tab:ablations}), starting from $M^{\mathcal{RL}}$ rather than $M_{\text{init}}$ is crucial for ProxyCoT-RL's performance.



\begin{table}[t]
\centering
\scalebox{0.85}{
\begin{tabular}{lrr}
\toprule
\textbf{Specification} & \textbf{Qwen3-4B} & \textbf{Gemma3-4B} \\
\midrule
Number of Parameters & 4B & 4B \\
Native Context Size & 256K & 32K \\
Supporting Context Size & 256K & 128K \\
Attention Architecture & Dense & Sparse \\
Release Date & July 2025 & March 2025 \\
\bottomrule
\end{tabular}
}
\caption{Details of experimental models: Qwen3-4B-Instruct-2507 and Gemma3-4B-IT. Gemma3-4B-IT supports the 128K-token context based on RoPE scaling.
\label{tab:model_details}} 
\end{table}

\section{Experimental Setup}

This section describes the models and datasets we use in our experiments. In addition to our experimental setting, we discuss implementation details and evaluation metrics.

\subsection{Models and Datasets}

We conduct experiments with two widely-used open-source long-context language models: Qwen3-4B-Instruct-2507~\citep{qwen3_technical_report_2025} and Gemma3-4B-IT~\citep{gemma3_technical_report_2025}.\footnote{We obtained models from \url{www.huggingface.co}: Qwen/Qwen3-4B-Instruct-2507 and google/gemma-3-4b-it. Qwen3-4B-Instruct abbreviates Qwen3-4B-Instruct-2507.} The two models differ substantially in architecture and modality, as shown in Table~\ref{tab:model_details}. Qwen3-4B-Instruct-2507 is dense and text-only, while Gemma3-4B-IT is sparse and multimodal.

We report results on two question answering benchmarks: SciTrek~\citep{scitrek_2025} and HotpotQA~\citep{hotpotqa_2018}. \textbf{SciTrek} is a long-context question answering benchmark, testing model reasoning capabilities over multiple scientific articles. Each instance consists of an \emph{article collection} formed by concatenating several articles to a target length (released at 64k, 128k, 512K, and 1M tokens), with collections constructed either by random sampling within topical clusters or by traversing citation graphs to preserve citation structures. SciTrek's questions generally require many steps of reasoning, such as comparison, sorting, filtering, and aggregating. 
As these questions (and their answers) are generated based on metadata about titles, authors, and references for each collection, we simply use a textual representation of the metadata as the proxy context of the full long context (see the example in Table~\ref{tab:example_proxy}).

\begin{table}[t]
\setlength{\tabcolsep}{2pt}
\centering
\scalebox{0.78}{
\begin{tabular}{@{}lp{9cm}@{}} \toprule
  \multicolumn{2}{c}{\bf SciTrek} \\ \toprule
$q$ & What is the smallest number of authors for any article in the collection?\\
\textbf{$C$} & \{7 full-text scientific articles\}\\
\textbf{$C^{p}$} & Article title: Existence and uniqueness for Legendre curves

There are 6 words in the title (separated by spaces).

There are 2 authors: Tomonori Fukunaga, Masatomo Takahashi

There are 9 references in the reference section.

The other provided articles are not cited by this article.

\textit{\{we cut down the text because of limited space.\}}

Article title: Effect of higher-order interactions on synchronization of neuron models with electromagnetic induction

There are 12 words in the title (separated by spaces).

There are 4 authors: Mohanasubha Ramasamy, Subhasri Devarajan, Suresh Kumarasamy, Karthikeyan Rajagopal

There are 37 references in the reference section.

The other provided articles are not cited by this article.\\ \bottomrule
\multicolumn{2}{c}{\bf HotpotQA} \\\toprule
\textbf{$q$} & Luis Gianneo was teacher of which chief exponent of Argentine folk music?\\
\textbf{$C$} & \{67 full-text Wikipedia articles\}\\
\textbf{$C^{p}$} & Luis Gianneo (1897–1968) was an Argentine composer, pianist and conductor.  

As music educator, he was the teacher of composers Ariel Ramirez, Juan Carlos Zorzi, Virtú Maragno, Pedro Ignacio Calderón and Rodolfo Arizaga, among others.\\
\bottomrule
\end{tabular}
}
\caption{Example questions ($q$), abbreviated long contexts($C$),
  and their proxy contexts ($C^{p}$) for SciTrek (top) and HotpotQA (bottom).} 
\label{tab:example_proxy}
\end{table}

\textbf{HotpotQA} is a multi-hop question answering benchmark based on Wikipedia articles, requiring models to reason and derive the answer from multiple documents. Each instance comprises a collection of Wikipedia articles and a question-answer pair. To create inputs of varying length up to 128K tokens, we extended each instance by appending additional full Wikipedia articles that are cited by the original context articles as needed. Details about how we construct long contexts for HotpotQA are provided in Appendix~\ref{sec:hotpotqa_construction}.  For each question, HotpotQA provides human-annotated supporting sentences, which are a subset of sentences from the corresponding input articles. We use the concatenation of these annotated evidence sentences as the proxy context, since they contain the information required to answer the question (see Table~\ref{tab:example_proxy}).

\begin{table}[t]
\centering
\setlength{\tabcolsep}{4pt}
\scalebox{0.85}{
\begin{tabular}{lcccc}
\toprule
\textbf{Dataset} & \textbf{Full Context} & \textbf{Proxy} & \textbf{Question} & \textbf{Answer}\\
\midrule
SciTrek & 83,018.3 & 659.4 & 18.9 & 16.7 \\
HotpotQA & 77,764.4 & 301.1 & 24.5 & ~~4.5 \\
\bottomrule
\end{tabular}
}
\caption{Dataset statistics for SciTrek and HotpotQA: average number of tokens in full contexts, proxy contexts, questions, and ground-truth answers, based on the tokenizer of Qwen3-4B-Instruct.}
\label{tab:datasets}
\end{table}


Due to limited computational resources, we restrict our experiments to data with contexts of up to~128K tokens (7,290/413/840 instances for SciTrek and 4,136/400/600 for HotpotQA in training/development/testing).\footnote{All experiments were conducted on 8 NVIDIA HGX H200 GPUs.} More statistics for SciTrek and HotpotQA are in Table~\ref{tab:datasets}.

\subsection{Implementation Details}

\paragraph{Large Teacher Sampling} We use Qwen3-235B-A22B-Thinking as the teacher model for both SciTrek and HotpotQA to generate reasoning traces on proxy contexts for Stage 1. It uses a Mixture-of-Experts architecture with 235B total parameters, of which only \textasciitilde22B are active during inference, and supports contexts of up to 256K tokens. Among the open-source models we evaluated, it achieved the strongest overall performance, motivating its use as our teacher model.  No training is required to collect reasoning traces from it; we sample three times and select only traces that result in correct answers. We run inference with vLLM~\citep{vllm_2023}, and temperature sampling set to~0.7, nucleus sampling with $top\_p = 0.8$, and top-$k$ sampling with $top\_k =20$. We set ${min\_p}$ to 0, and cap generations to a maximum of~32,768 tokens.

\paragraph{RLVR Training} For RL training in Stage 1 (see Figure~\ref{fig:proxy_pipeline}), we optimize our models using DAPO~\citep{dapo_2025}. At each training step, DAPO samples a group of reasoning traces from the policy and computes advantages for each trace through group-based normalization of their rewards. The model parameters are then updated based on these advantage estimates. As both of our tasks require short strings as answers, we adopt a simple sum of F1 and exact match to represent the reward as a function of the ground-truth answer $a$ and our predicted answer $\hat{a}$:
\begin{equation}
  R(a,\hat{a}) = \text{F1}(a,\hat{a}) + \mathds{1}_{a==\hat{a}} 
\end{equation}
We perform RL training with OpenRLHF~\citep{hu2024openrlhf} with a batch size of~64 and a maximum generation length of~2,048 tokens. We set the actor learning rate to~5e-7, apply dynamic reward filtering with a range of (0.3, 2.0), and use clipping parameters (0.2, 0.3) for the policy update. For each prompt, we sample 8 trajectories. Models are trained for 10 epochs.


\paragraph{SFT on CoTs} In Stage~2,  both ProxyCoT-ZS and ProxyCoT-RL use supervised fine-tuning (SFT) to distil chain-of-thought traces. We implement SFT with OpenRLHF~\citep{hu2024openrlhf}, using a batch size of 64 and a learning rate of 5e-6.  We apply linear learning-rate warm up over the first 10\% of training steps. 

\begin{table}[t]
\centering
\setlength{\tabcolsep}{1.5pt}
\scalebox{0.85}{
\begin{tabular}{@{}l@{~}c@{~}c@{~}c@{}}
\toprule
\multicolumn{1}{c}{\textbf{Model}} & \multicolumn{1}{c}{\textbf{Training Strategy}} & \textbf{Proxy$\uparrow$} & \textbf{Full$\uparrow$}\\
\midrule
Qwen3-235B-Instruct  & Zero-shot & 72.7 & 45.2 \\
Qwen3-235B-Thinking  & Zero-shot & 85.6 & 48.8 \\
\midrule
\multirow{6}{3cm}{Qwen3-4B-Instruct}  & Zero-shot & 67.2 & 30.8 \\
& SFT on $C$ & 39.0 & 19.5 \\
& RLVR on $C$ & 66.1 & 32.9 \\
& SFT on $C$, CoT$^*$ & 45.8 & 31.6 \\
& ProxyCoT-ZS  & 67.8 & 38.8 \\
& ProxyCoT-RL & 88.5 & 46.5\\
\midrule
\multirow{6}{3cm}{Gemma3-4B-IT} & Zero-shot & 34.2 & ~~3.0 \\
& SFT on $C$ & 19.1 & 12.7\\
& RLVR on $C$ & 39.9 & ~~5.5 \\
& SFT on $C$, CoT$^*$ & 53.1 & 36.9 \\
& ProxyCoT-ZS  & 64.2 & 36.5 \\
& ProxyCoT-RL  & 69.8 & 43.7 \\
\bottomrule
\end{tabular}
}
\caption{Performance of Qwen3-4B-Instruct and Gemma3-4B-IT on SciTrek, in terms of exact match~(\%). Models are evaluated with Proxy and Full contexts as input. $C$: full long contexts, and CoT$^*$: chain-of-thought reasoning traces  generated by Qwen3-235B-A22B-Thinking on full long contexts.} 
\label{tab:model_performances_scitrek}
\end{table}

\subsection{Evaluation Metrics}

We generate final answers using the default decoding settings for Qwen3-4B-Instruct and Gemma3-4B-IT. For models based on Qwen3-4B-Instruct, we decode with temperature~0.7, $top\_p=0.8$, ${top\_k=20}$, ${min\_p= 0}$, and a maximum generation length of 2,048 tokens. For Gemma3-4B-IT based models, we use temperature 1.0, ${top\_p=0.95}$, ${top\_k}=64$, and a maximum of 2,048 generated tokens.

We evaluate model performance based on the quality of the generated answers. For SciTrek, following~\citet{scitrek_2025}, we use exact match and F1 comparing the generated answer against the ground-truth answer. For HotpotQA, where acceptable answers can be more variable, we adopt a model-based evaluation protocol following previous work~\cite{sun-etal-2024-head, uncertainty_quantification_2025}. We use GPT5-mini as the judge, assessing the generated answer against the ground-truth answer in the prompt. The evaluation prompt and more details are provided in Appendix~\ref{sec:model_as_judge}.

\begin{table}[t]
\centering
\setlength{\tabcolsep}{1.5pt}
\scalebox{0.85}{
\begin{tabular}{@{}l@{~}c@{~}cc@{}}
\toprule
\multicolumn{1}{c}{\textbf{Model}} & \textbf{Training Strategy} & \textbf{Proxy$\uparrow$} & \textbf{Full$\uparrow$}\\
\midrule
Qwen3-235B-Instruct  & Zero-shot & 92.1 & 60.8 \\
Qwen3-235B-Thinking  & Zero-shot & 93.2 & 50.7 \\
\midrule
\multirow{6}{3cm}{Qwen3-4B-Instruct} & Zero-shot & 91.3 & 44.5 \\
& SFT on $C$ & 92.6 & 48.8 \\
& RLVR on $C$ & 88.6 & 48.1 \\
& SFT on $C$, CoT$^*$ & 84.5 & 40.2\\
& ProxyCoT-ZS & 91.4 & 50.3 \\
& ProxyCoT-RL & 92.1 & 52.7 \\
\bottomrule
\end{tabular}
}
\caption{Performance of Qwen3-4B-Instruct on HotpotQA evaluated with GPT5-mini as judge in terms of accuracy (\%). Models are evaluated with Proxy and Full contexts as input. $C$: full long contexts, and CoT$^*$: chain-of-thought reasoning traces from Qwen3-235B-A22B-Thinking on full long contexts.}
\label{tab:model_performances_qwen3_hotpotqa}
\end{table}

\section{Main Results}
 
In this section, we present experimental results comparing our
training framework against a range of baselines. We further analyze
the reasoning traces produced by each method and evaluate robustness
via an out-of-domain transfer task.

\paragraph{ProxyCoT improves long-context reasoning across model families.} We compare our training framework with several baselines, which are based on Qwen3-4B-Instruct or Gemma3-4B-IT and do not make use of proxy contexts. These baselines include: (1) supervised fine-tuning (SFT) on full long contexts without generating reasoning traces, (2) reinforcement learning with verifiable rewards (RLVR) on full contexts, and (3) SFT on full long contexts using reasoning traces generated by Qwen3-235B-A22B-Thinking. As an upper bound, we also compare the smaller models against zero-shot Qwen3-235B-A22B-Thinking. 

Results in Table~\ref{tab:model_performances_scitrek} show that ProxyCoT-ZS and ProxyCoT-RL consistently outperform these training alternatives, improving performance on both full long contexts and short proxy contexts, and across model architectures. Notably, ProxyCoT-RL substantially improves Qwen3-4B-Instruct, achieving performance competitive with the much larger Qwen3-235B-A22B-Thinking model. Moreover, ProxyCoT-RL consistently outperforms ProxyCoT-ZS, suggesting that reasoning traces generated via RLVR can be more effective distillation targets than zero-shot traces from a large teacher model.
It is worth noting that operating on short proxy contexts is equivalent to retrieval-augmented generation (RAG) with perfect retrieval, i.e., an oracle retriever that always selects exactly the right evidence, thereby representing an upper bound on what any RAG approach can achieve (the second-to-last column in Table~\ref{tab:model_performances_scitrek}).


\paragraph{ProxyCoT performs well across datasets.} Table~\ref{tab:model_performances_qwen3_hotpotqa} reports results on HotpotQA, corroborating the trends observed on SciTrek: ProxyCoT consistently improves long-context question answering. Although the baseline Qwen3-4B-Instruct model is already strong on proxy inputs, ProxyCoT yields clear gains when evaluated on full long contexts, outperforming both SFT and RLVR on full contexts. In particular, ProxyCoT-RL achieves the best performance on full contexts, indicating that distilling high-quality traces obtained on proxy inputs remains effective for improving long-context reasoning on HotpotQA.


\paragraph{ProxyCoT-RL reduces inference compute.} Beyond eliminating the need to generate teacher traces on full long contexts and requiring less computation during reinforcement learning, ProxyCoT-RL also induces substantially shorter reasoning traces at inference time. As shown in Table~\ref{tab:inference_budget} (Qwen3-4B-Instruct on SciTrek), ProxyCoT-RL uses fewer CoT tokens on average while achieving better accuracy than the other optimization strategies. These results suggest that ProxyCoT-RL improves long-context performance while simultaneously reducing inference-time budget.

\begin{table}[t]
\setlength{\tabcolsep}{4pt}
\centering
\scalebox{0.85}{
\begin{tabular}{lcc}
\toprule
\textbf{Training Strategy} & \textbf{CoT Tokens$\downarrow$} & \textbf{EM$\uparrow$} \\
\midrule
Zero-shot & 1,744 & 30.8 \\
RLVR on $C$ & ~~~937 & 32.9\\
SFT on $C$, CoT$^*$ & 6,683 & 31.6 \\
ProxyCoT-ZS  & 5,520 & 38.8\\
ProxyCoT-RL & ~~~617 & 46.5\\
\bottomrule
\end{tabular}
}
\caption{Models trained with different training strategies use different numbers of chain-of-thought (CoT) tokens when showing full long contexts. Results are based on Qwen3-4B-Instruct on SciTrek. CoT$^*$: chain-of-thought reasoning traces are generated by Qwen3-235B-A22B-Thinking on full long contexts.} 
\label{tab:inference_budget}
\end{table}

\paragraph{ProxyCoT-RL also pushes out-of-domain long-context capabilities.}

To further validate the out-of-domain generalization capabilities of ProxyCoT, we report results on Loong~\citep{loong_2024}, a benchmark designed to evaluate long-context language models through extended multi-document question answering (every document in each test case must be considered to derive the final answer). Loong features four long-context task types: Spotlight Locating, Comparison, Clustering, and Chain of Reasoning, and represents the domains of Financial Reports, Academic Papers, and Legal Cases (with context lengths ranging from 10K to beyond 200K tokens). We evaluate ProxyCoT-RL (trained on SciTrek) on Loong, considering English instances only up to 128K tokens without any further adaptation.\footnote{We tested our models on 309 instances in Financial Reports and 119 in Academic Papers, omitting instances in the domain of Legal Cases which are all in Chinese.}

As shown in Table~\ref{tab:loong_performance}, both Qwen3-4B-Instruct and Gemma3-4B-IT demonstrate improved performance on Loong across different domains after training with ProxyCoT-RL.\footnote{We follow the evaluation prompt and code from \url{https://github.com/MozerWang/Loong}.} On financial reports, both models show gains, with Gemma3-4B-IT exhibiting particularly substantial improvement (25.85 $\rightarrow$ 32.05). More notably, on academic papers, a domain represented in Loong but with different questions than those in SciTrek, both models achieve substantial performance increase, with Gemma3-4B-IT improving from 3.55 to 24.32. These cross-domain gains demonstrate that ProxyCoT-RL enhances general long-context reasoning capabilities of the models rather than simply memorizing task-specific patterns.


\begin{table}[t]
\setlength{\tabcolsep}{1.5pt}
\centering
\scalebox{0.83}{
\begin{tabular}{lccc}
\toprule
\multicolumn{1}{c}{\textbf{Model}} & \textbf{Training} & \textbf{Financial$\uparrow$} & \textbf{Academic$\uparrow$} \\
\midrule
\multirow{2}{3cm}{Qwen3-4B-Instruct} & Zero-shot & 37.76 & 24.91 \\
 & ProxyCoT-RL & 40.83 & 42.51 \\
\midrule
\multirow{2}{3cm}{Gemma3-4B-IT} & Zero-shot & 25.85 & ~~3.55\\
 & ProxyCoT-RL & 32.05 & 24.32 \\
\bottomrule
\end{tabular}
}
\caption{Results on Loong (domains of Financial Reports and Academic Papers) when the model is zero-shot or trained on SciTrek with ProxyCoT-RL, in terms of answer quality (scored on a 1–100 scale by GPT-5 mini as the judge).}
\label{tab:loong_performance}
\end{table}

\section{Analysis and Ablations}

We further analyze the significance of the two training stages in our framework, and conduct experiments with alternative proxy contexts to examine whether their quality impacts performance.


\subsection{Ablations on Two-stage Training}


Table~\ref{tab:ablations} summarizes ablation studies with both Qwen3-4B-Instruct and Gemma3-4B-IT. We report performance on proxy and full contexts in three settings: (a) our full training with both RLVR and SFT; (b)~only RLVR-based training; and (c)~only SFT-based training but with reasoning traces from RLVR. Models trained with RLVR learn to reason based on proxy contexts and can be applied to long contexts without any grounding. Models trained with SFT learn the task by observing long contexts, questions, and reasoning traces with answers.


\begin{table}[t]
\centering
\setlength{\tabcolsep}{4pt}
\scalebox{0.85}{
\begin{tabular}{@{}lccccc@{}}
\toprule
& \textbf{Stage 1} & \textbf{Stage 2} &  & \\
\multicolumn{1}{c}{\textbf{Model}} & \textbf{(RLVR)} & \textbf{(SFT)} & \textbf{Proxy$\uparrow$} & \textbf{Full$\uparrow$}\\
\midrule
\multirow{3}{3cm}{Qwen3-4B-Instruct} & \cmark & \cmark & 88.5 & 46.5 \\
& \cmark & \xmark & 91.5 & 29.0\\
& \xmark & \cmark & 77.5 & 46.3 \\

\midrule
\multirow{3}{3cm}{Gemma3-4B-IT} & \cmark & \cmark & 69.8 & 43.7 \\
& \cmark & \xmark & 88.5 & ~~8.0 \\
& \xmark & \cmark & 65.2 & 37.3 \\

\bottomrule
\end{tabular}
}
\caption{Ablations on ProxyCoT-RL using Qwen3-4B-Instruct and Gemma3-4B-IT with SciTrek. We report results in terms of exact match (\%) on proxy and full contexts. There are two stages in training: reinforcement learning with verifiable rewards (RLVR) in Stage~1, and supervised fine-tuning (SFT) in Stage~2.}
\label{tab:ablations}
\end{table}

Our results show model performance on full long contexts is best when \emph{both} training stages are employed (see column Full in Table~\ref{tab:ablations}). When only reinforcement learning takes place (Stage 1), we observe that performance on proxy contexts is superior to the other two strategies (i.e.,~Stage 2 or Stage 1 + Stage 2). We also find the first stage of training gives Gemma3-4B-IT a larger performance boost in long contexts (37.3 $\rightarrow$ 43.7) than Qwen3-4B-Instruct (46.3 $\rightarrow$ 46.5). As Gemma3-4B-IT has worse zero-shot performance than Qwen3-4B-Instruct (see Table~\ref{tab:model_performances_scitrek}), we conjecture that models with poorer zero-shot capabilities at long contexts may benefit more from RL on proxy contexts. 



\subsection{Alternative Proxy Contexts}
\label{sec:alternative-proxies}


Proxy contexts in our experiments were obtained from annotations provided in SciTrek and HotpotQA (see examples in Table~\ref{tab:example_proxy}). In this section, we first experiment with automatically created proxies under the assumption that high-quality annotations might be scarce.
We examine different proxies for full long contexts based on information retrieval. 

For SciTrek, we simply use article titles, authors, and reference sections of input articles as the proxy context, since the questions (and their answers) focus on information contained therein. For HotpotQA, we use sentences semantically related to the question as the proxy context, assuming that unrelated sentences introduce noise into the answer generation process. We first obtain related sentences via lexical search with BM25~\citep{bm25_2009} and then via semantic search based on sentence embeddings (\textit{text-embedding-3-small} from OpenAI). Finally, we retrieve random sentences as proxy contexts for both datasets. 

\begin{table}[t]
\centering
\setlength{\tabcolsep}{3.5pt}
\scalebox{0.85}{
\begin{tabular}{@{}l@{~}p{5.5cm}c@{~}c@{}}
  \toprule
&  \multicolumn{1}{c}{\textbf{Proxy Type}} & \textbf{\#Tokens} & \textbf{Perf.$\uparrow$}\\
\midrule
& Random sentences from full context & ~1,102 & ~~3.4 \\
& Titles, authors and references of the context articles & 20,344 & 24.6 \\
\raisebox{.05cm}[0pt]{\begin{sideways}\textbf{SciTrek}\end{sideways}} & Descriptions of the structured metadata (provided by SciTrek) & ~~~~659 & 91.5 \\\midrule
& Random sentences from  full context & ~~~~385  & 15.8 \\
& Sentences semantically related to the question & ~~~~364 & 35.1 \\
\raisebox{-.1cm}[0pt]{\begin{sideways}\textbf{HotpotQA}\end{sideways}}&
Ground-truth evidence sentences (provided by HotpotQA) & ~~~~301 & 92.2 \\
\bottomrule
\end{tabular}
}
\caption{Performance of Qwen3-4B-Instruct after RLVR with different
  types of proxy contexts on SciTrek and HotpotQA in terms of exact match (\%) and accuracy (\%, evaluated with GPT5-mini as judge), respectively.}
\label{tab:proxies}
\end{table}

As shown in Table~\ref{tab:proxies}, proxy contexts based on annotations are of much higher quality compared to automatically sourced ones. Perhaps unsurprisingly, proxies based on randomly selected sentences carry no useful information for answering the question. For HotpotQA, semantically related sentences based on information retrieval are better than random ones but still a poor substitute for annotation-based proxies. For SciTrek, even though titles, authors and references are the only relevant pieces of information for answering the question, their unstructured nature makes learning difficult.  Structured metadata is far more advantageous for learning targeted reasoning traces than pure text (despite the two formats being equally sufficient to answer the question). This analysis shows that better designed proxy contexts enable ProxyCoT to improve long-context reasoning.

We further evaluate the robustness of ProxyCoT-RL by introducing controlled noise into the proxy contexts. Specifically, we train Qwen3-4B-Instruct using our ProxyCoT-RL on both SciTrek and HotpotQA with degraded proxies. These proxies are constructed by augmenting the oracle evidence (i.e., the human-annotated proxy contexts) with randomly injected irrelevant sentences. The results in Table~\ref{tab:noise_proxies} demonstrate that ProxyCoT-RL remains robust under noisier conditions, with performance degrading only marginally as the proportion of injected noise increases.

\begin{table}[t]
\centering
\setlength{\tabcolsep}{4pt}
\scalebox{0.9}{
\begin{tabular}{lcc@{~}c}
  \toprule
\textbf{Dataset} & \textbf{Oracle: Noise} & \textbf{\#Tokens} & \textbf{Performance$\uparrow$}\\
\midrule
\multirow{4}{*}{SciTrek} & 1:5 & 5,471 & 85.3 \\
& 1:2 & 2,588 & 89.9 \\
& 1:1 & 1,622 & 91.3 \\
& 1:0 & ~~~659 & 91.5 \\
\midrule
\multirow{4}{*}{HotpotQA} & 1:5 & 1,853 & 83.7 \\
& 1:2 & ~~~922 & 88.4 \\
& 1:1 & ~~~607 & 90.1 \\
& 1:0 & ~~~301 & 92.2 \\
\bottomrule
\end{tabular}
}
\caption{Performance of Qwen3-4B-Instruct trained with ProxyCoT-RL on noisy proxy contexts:  human-annotated proxies augmented with randomly sampled sentences as noise. Results on SciTrek are reported as exact match (\%); on HotpotQA as accuracy (\%), judged by GPT5-mini. The Oracle:Noise column gives the ratio of oracle evidence sentences to injected noise sentences; 1:0 denotes no injected noise.}
\label{tab:noise_proxies}
\end{table}




\section{Conclusion}
\label{sec:conclusion}

We introduced ProxyCoT, a two-stage training framework for long-context reasoning that leverages \emph{proxy contexts}, i.e.,~shorter inputs that preserve the information needed to solve a task. ProxyCoT is motivated by a consistent performance gap between full long contexts and their proxy counterparts, despite requiring the same underlying reasoning. Across model families and benchmarks, we show that ProxyCoT improves long-context question answering while reducing inference-time budget. More broadly, ProxyCoT offers a practical route to strengthening long-context reasoning without relying on teacher-generated traces over full long contexts, providing an efficient alternative in settings where training and supervising long-context models remain challenging.

\section*{Limitations}
While learning via proxy contexts offers a promising direction for improving long-context reasoning, our work has several limitations:
\begin{itemize}
\item Our approach assumes access to proxy contexts that contain sufficient evidence to answer each question. Constructing such proxies can be non-trivial for some real-world tasks and domains. In future work, we plan to explore methods for automatically constructing effective proxy contexts.

\item  Some of our tasks can also be addressed with workflow-based solutions, such as retrieval-augmented generation, which retrieves relevant evidence and conditions generation on the retrieved subset. Studying such systems-level approaches is outside the scope of this paper; our focus is on improving the inherent long-context capabilities of language models.

\item Although our framework generalizes across multiple model families and datasets, our experiments are limited to English due to data availability and computational constraints. Extending to other languages and domains is an important direction for future work, contingent on obtaining suitable proxy contexts.
\end{itemize}

 \section*{Acknowledgments}
We thank the meta-reviewer and anonymous reviewers for their constructive feedback. We gratefully acknowledge the support of the UK Engineering and Physical Sciences Research Council (grant EP/W002876/1) and the support of the Edinburgh International Data Facility (EIDF) and the Data-Driven Innovation Programme at the University of Edinburgh. 

\bibliography{references}

\clearpage

\onecolumn
\appendix
\section{Long Context Construction for HotpotQA}
\label{sec:hotpotqa_construction}

For the long contexts of HotpotQA, we retained the original questions and answers from~\citet{hotpotqa_2018} and constructed long contexts using the procedure described by \citet{scitrek_2025}.  We used the distractor setting and selected only bridge questions at the hard difficulty level. Bridge questions require chain reasoning, where the model must first identify a bridge entity connecting two paragraphs before completing the second reasoning step. We further filtered for examples with at least 3 supporting facts, i.e., sentences that crowd workers annotated as necessary for answering the question. Our training and validation sets were sampled from the original train split, while our test set was sampled from the original validation split, resulting in 4136 training examples, 400 validation examples, and 600 test examples.

We first retrieved the full texts of the Wikipedia articles containing the supporting facts, as well as the articles corresponding to the retrieved paragraphs in the distractor setting. We used the preprocessed Wikipedia dump released with HotpotQA for the retrieval of all articles. We then expanded the context to 128K tokens by adding related Wikipedia articles: we gathered the links mentioned in each article and followed them to retrieve additional articles. We limited the maximum depth to two hops from the original Wikipedia articles. Finally, we concatenated all articles to form the full long context. 

\section{Evaluation Metrics with Model-as-Judge for HotpotQA}
\label{sec:model_as_judge}

To compare predicted and ground-truth answers on HotpotQA, we use the prompt from~\citet{sun-etal-2024-head}, who found 98\% agreement between human judgments and LLM-based scores. We use GPT-5-mini as a judge (\texttt{gpt-5-mini-2025-08-07}, \url{https://platform.openai.com/docs/models/gpt-5-mini}). The prompt template based on few-shot prompting obtained from~\citet{sun-etal-2024-head} is given below. 
\bigskip

  \begin{tcolorbox}[breakable,colback=gray!20!white,colframe=NavyBlue,title=The Prompt Template for HotpotQA Evaluation,fonttitle=\bfseries, halign title=flush center]
    You need to check whether the prediction of a question-answering system to a question is correct.\\
    You should make the judgment based on a list of ground truth answers provided to you.\\
    Your response should be "correct" if the prediction is correct or "incorrect" if the prediction is wrong.\\

    Question: Who authored The Taming of the Shrew (published in 2002)? \\
    Ground truth: ["William Shakespeare", "Roma Gill"] \\
    Prediction: W Shakespeare \\
    Correctness: correct \\
    
    Question: Who authored The Taming of the Shrew (published in 2002)?\\
    Ground truth: ["William Shakespeare", "Roma Gill"]\\
    Prediction: Roma Gill and W Shakespeare\\
    Correctness: correct\\
    
    Question: Who authored The Taming of the Shrew (published in 2002)?\\
    Ground truth: ["William Shakespeare", "Roma Gill"]\\
    Prediction: Roma Shakespeare\\
    Correctness: incorrect\\
    
    Question: What country is Maharashtra Metro Rail Corporation Limited located in?\\
    Ground truth: ["India"]\\
    Prediction: Maharashtra\\
    Correctness: incorrect\\
    
    Question: What's the job of Song Kang-ho in Parasite (2019)?\\
    Ground truth: ["actor"]\\
    Prediction: He plays the role of Kim Ki-taek, the patriarch of the Kim family.\\
    Correctness: correct\\
    
    Question: Which era did Michael Oakeshott belong to?\\
    Ground truth: ["20th-century philosophy"]\\
    Prediction: 20th century.\\
    Correctness: correct\\
    
    Question: Edward Tise (known for Full Metal Jacket (1987)) is in what department?\\
    Ground truth: ["sound department"]\\
    Prediction: 2nd Infantry Division, United States Army\\
    Correctness: incorrect\\
    
    Question: What wine region is Finger Lakes AVA a part of?\\
    Ground truth: ["New York wine"]\\
    Prediction: Finger Lakes AVA\\
    Correctness: incorrect\\
    
    Question: $\{$QUESTION$\}$\\
    Ground truth: $\{$GROUND\_TRUTH$\}$\\
    Prediction: $\{$PREDICTION$\}$\\
    Correctness:
    
    \end{tcolorbox}
    \label{fig:prompt_hotpotqa_eval}

\end{document}